\documentclass[letterpaper]{article} 
\usepackage{aaai2026}  
\usepackage{times}  
\usepackage{helvet}  
\usepackage{courier}  
\usepackage[hyphens]{url}  
\usepackage{graphicx} 
\usepackage{natbib}  
\usepackage{caption} 
\usepackage{algorithm}
\usepackage{algorithmic}

\usepackage{newfloat}
\usepackage{listings}
\DeclareCaptionStyle{ruled}{labelfont=normalfont,labelsep=colon,strut=off} 
\floatstyle{ruled}
\newfloat{listing}{tb}{lst}{}
\floatname{listing}{Listing}
\usepackage{amsmath,amssymb}
\usepackage{booktabs,multirow}
\title{TripleFDS: Triple Feature Disentanglement and Synthesis for Scene Text Editing}
\author {
    Yuchen Bao\textsuperscript{\rm 1, 2}\thanks{Work done during internship at Tencent Youtu Lab.},
    Yiting Wang\textsuperscript{\rm 2},
    Wenjian Huang\textsuperscript{\rm 1},
    Haowei Wang\textsuperscript{\rm 2},
    Shen Chen\textsuperscript{\rm 2},
    Taiping Yao\textsuperscript{\rm 2},
    Shouhong Ding\textsuperscript{\rm 2},
    Jianguo Zhang\textsuperscript{\rm 1, 3}\thanks{Corresponding author.}
}
\affiliations {
    \textsuperscript{\rm 1}Research Institute of Trustworthy Autonomous Systems and Department of Computer Science and Engineering, Southern University of Science and Technology, Shenzhen 518055, China\\
    \textsuperscript{\rm 2}Tencent Youtu Lab
    \textsuperscript{\rm 3}Guangdong Provincial Key Laboratory of Brain-inspired Intelligent Computation, Department of Computer Science and Engineering, Southern University of Science and Technology, Shenzhen, China
}

\usepackage{bibentry}

\begin{document}

\maketitle

\begin{abstract}
Scene Text Editing (STE) aims to naturally modify text in images while preserving visual consistency, the decisive factors of which can be divided into three parts, \textit{i.e.}, \textit{text style}, \textit{text content}, and \textit{background}. Previous methods have struggled with \textit{incomplete disentanglement of editable attributes}, typically addressing only one aspect—such as editing text content—thus limiting controllability and visual consistency. To overcome these limitations, we propose \textbf{TripleFDS}, a novel framework for STE with disentangled modular attributes, and an accompanying dataset called \textbf{SCB Synthesis}. 
\textbf{SCB Synthesis} provides robust training data for triple feature disentanglement by utilizing the ``SCB Group", a novel construct that combines three attributes per image to generate diverse, disentangled training groups.
Leveraging this construct as a basic training unit, \textbf{TripleFDS} first disentangles triple features, ensuring semantic accuracy through inter-group contrastive regularization and reducing redundancy through intra-sample multi-feature orthogonality. In the synthesis phase, \textbf{TripleFDS} performs feature remapping to prevent ``shortcut'' phenomena during reconstruction and mitigate potential feature leakage.
Trained on 125,000 SCB Groups, \textbf{TripleFDS} achieves state-of-the-art image fidelity (SSIM of 44.54) and text accuracy (ACC of 93.58\%) on the mainstream STE benchmarks. Besides superior performance, the more flexible editing of \textbf{TripleFDS} supports new operations such as style replacement and background transfer. Code: \textcolor{blue}{https://github.com/yusenbao01/TripleFDS}
\end{abstract}
\vspace{0.5cm}

\section{Introduction}
\begin{figure}[t]
\centering
\includegraphics[width=1.0\columnwidth]{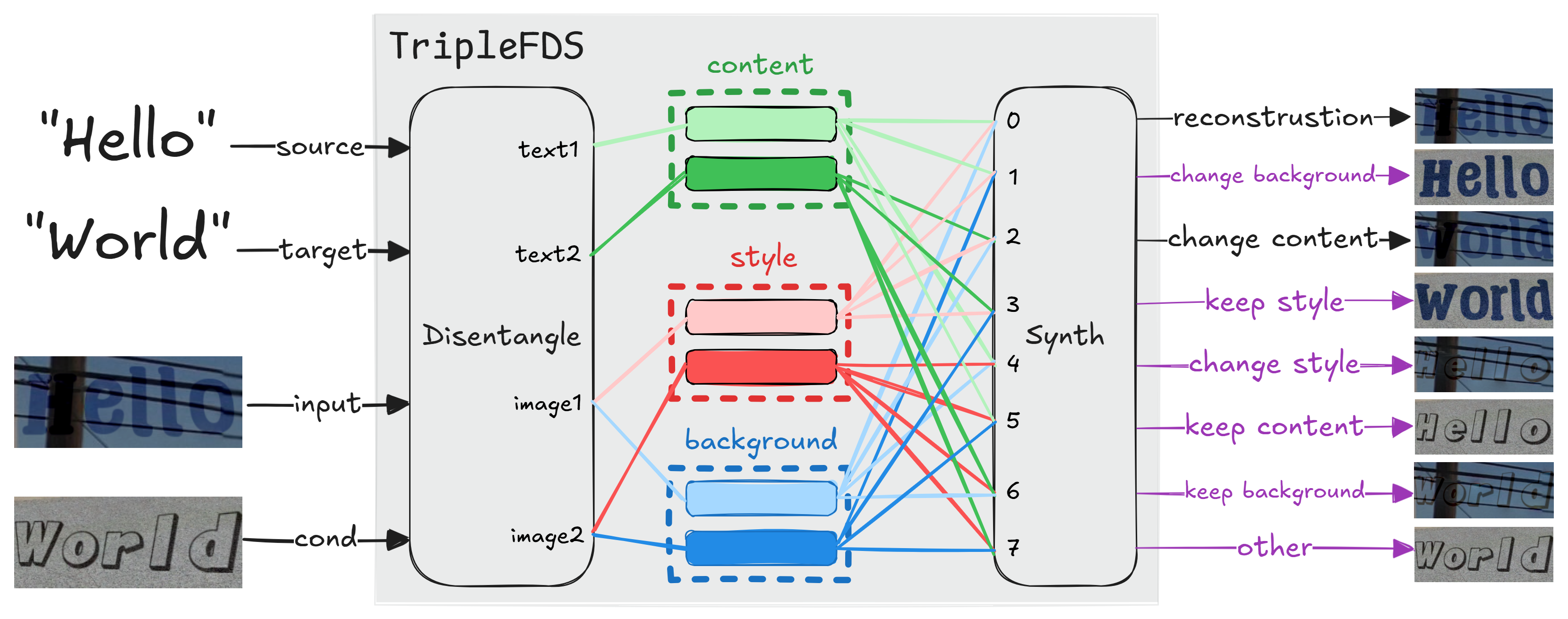}
\caption{\textbf{TripleFDS}'s capabilities. Purple lines denote additional feature permutation-based editing operations enabled by our approach.}
\label{fig:tasks}
\end{figure}

Scene Text Editing (STE) empowers users with fine-grained control to modify specific text within an image while preserving the invariance of other elements. As exemplified in Fig.~\ref{fig:tasks}, this technology is crucial for diverse applications, including digital content creation (e.g., editing documents and posters) and enhancing various downstream tasks such as Scene Text Removal \cite{darling}, Optical Character Recognition (OCR) \cite{rsste}, and Anti-forgery \cite{tamper1, tamper2, tamper3, tamper4, zhang2023toward}.

Scene Text Editing (STE) methods often employ non-disentangling or implicit strategies, such as diffusion-based inpainting methods \cite{textdiffuser,anytext,diffute} and Transformer-decoder methods \cite{darling,rsste}, enabling text reconstruction and inpainting with notable quality. Early attempts at explicit disentanglement, predominantly GAN-based \cite{synthtext, srnet,mostel,swaptext,textstylebrush,gan}, used distinct modules to separate visual components for tasks like text conversion and style/background extraction. More advanced explicit methods, such as diffusion-based reconstruction paradigms \cite{scenevtg,textctrl,fontdiffuser}, brought enhanced image quality and control by focusing on rich font feature representation and integrating noisy-latent via conditional controls (e.g., ControlNet\cite{contronet} for glyphs\cite{anytext2, glyphcontrol}, DINOv2\cite{dinov2} for styles\cite{textmaster, glyphbyt5} through bypass branches or cross-attention mechanisms\cite{ipadapter, improving}).

Despite these advances, existing STE methods are primarily limited by \textit{incomplete disentanglement of editable attributes}. For example, RS-STE \cite{rsste}'s insufficient disentanglement often leads to artifacts and blurriness, especially with elaborate fonts or complex backgrounds. Similarly, TextCtrl\cite{textctrl} employs binary disentanglement, where text content is separated, but style and background remain entangled. This results in inherent stylistic entanglement, causing style deviation and unnatural foreground-background boundaries.

To overcome the significant challenge of \textit{incomplete disentanglement of editable attributes} in previous methods, we propose \textbf{TripleFDS}, a novel framework for STE that explicitly disentangles triple features through a dedicated disentanglement and synthesis process. To support this, we introduce \textbf{SCB Synthesis}, a novel paradigm for constructing synthetic text image datasets. 

To achieve robust triple feature disentanglement, \textbf{SCB Synthesis} utilizes the core concept of the SCB Group, which facilitates the natural blending of permutations of text styles, text contents, and backgrounds, generating diverse and well-disentangled training samples. Furthermore, the model can perform self-supervision for features lacking explicit ground truth labels, such as style, by leveraging fixed mapping relationships among samples within a single SCB Group.

Building upon this data construct, we propose two distinct processes for feature disentanglement and synthesis. In the first process, we introduce novel feature disentanglement constraints to achieve accurate and non-redundant feature disentanglement by leveraging the mapping properties both within and across SCB Groups. This strategy enhances feature semantic accuracy, improves feature localization, and enforces orthogonality among features to reduce information redundancy.

In the synthesis process, unlike previous methods that primarily focus on editing, our approach emphasizes robust reconstruction. To achieve this, we introduce a feature remapping strategy that prevents ``shortcut" phenomena during reconstruction and mitigates potential feature leakage. Leveraging the fixed feature mapping relationships within the SCB Group, we deliberately create ``hard-to-reconstruct" inputs for image A, whose triple features are synthesized by remapping from other images (B, C, and D), compelling the model to generate purer triple features.

In summary, our main contributions are as follows:
\begin{itemize}
\item We propose \textbf{TripleFDS}, a novel framework combining explicit disentanglement and synthesis. It achieves accurate feature disentanglement through a self-supervised regularization strategy and ensures robust synthesis via feature remapping.
\item We introduce a new dataset, \textbf{SCB Synthesis}. By leveraging SCB Groups, it enhances training data diversity and facilitates robust disentanglement.
\item \textbf{TripleFDS} achieves state-of-the-art performance on mainstream STE benchmarks, while enabling flexible triple feature combinations for new operations like style replacement and background transfer.
\end{itemize}

\begin{figure}[t]
\centering
\includegraphics[width=1.0\columnwidth]{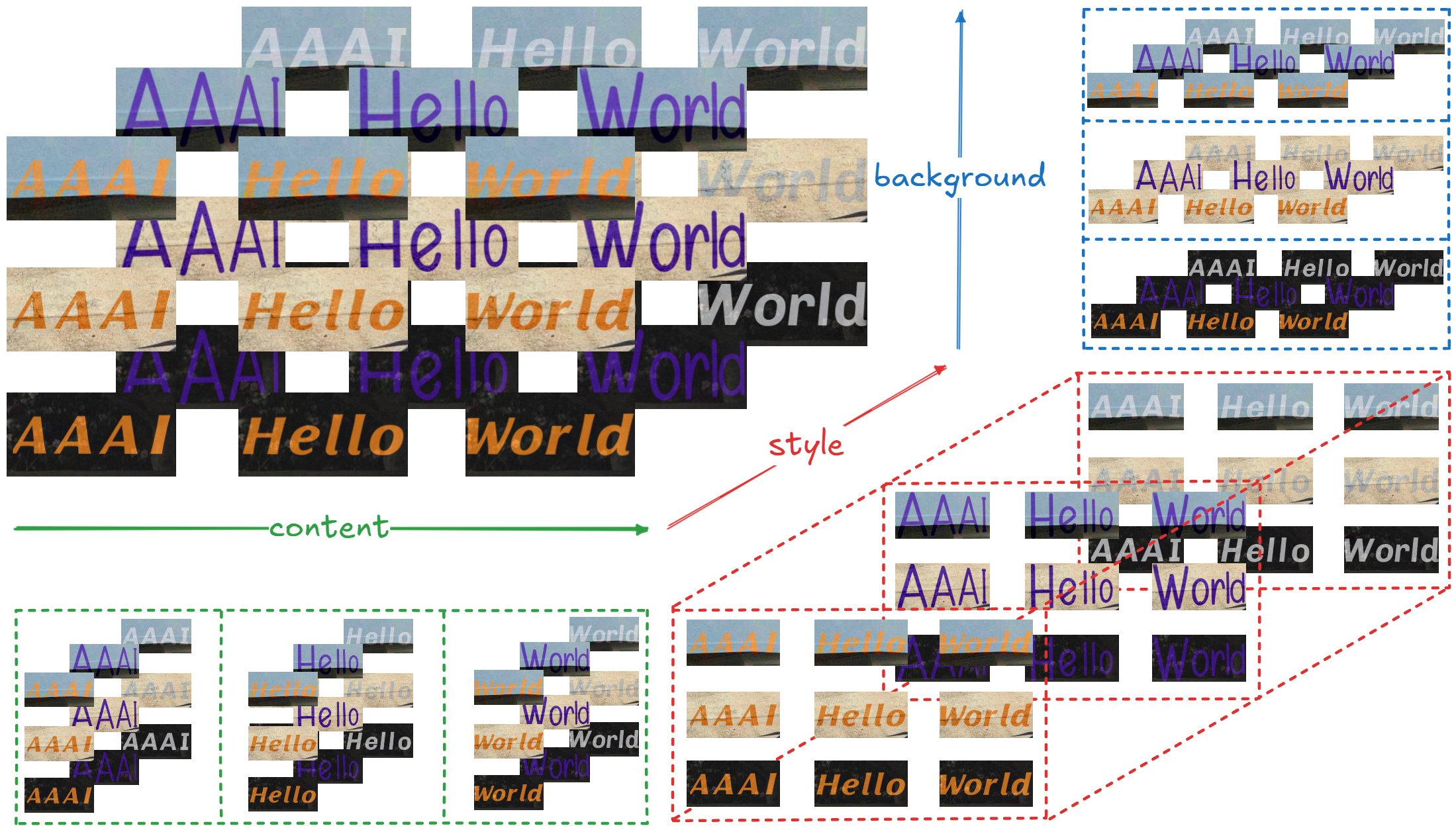} 
\caption{Visualizing $3\times3\times3$ SCB Group and Feature Disentanglement.}
\label{fig:scb} 
\end{figure}

\section{Related Work}
Existing techniques for Scene Text Editing (STE) can be broadly categorized based on their approach to feature disentanglement: non-disentangling/implicit editing, and explicit disentanglement.

\subsection{Non-disentangling or Implicit Editing Methods}
These methods primarily focus on text reconstruction or inpainting, without explicitly disentangling the aftermentioned triple features.

Diffusion-inpainting-based methods, such as TextDiffuser \cite{textdiffuser}, AnyText \cite{anytext}, Brush Your Text \cite{brushyourtext}, UDiffText \cite{udifftext}, DreamText \cite{dreamtext}, and Type-R \cite{typer}, perform local text erasure and masked region filling. Despite advancements in quality, their implicit handling of features often leads to stylistic entanglement, which results in blurriness caused by residuals.

Transformer-decoder-based methods, such as DARLING \cite{darling} and RS-STE \cite{rsste}, leverage sequence-to-sequence generation and attention\cite{attention} mechanisms for text editing. While achieving text consistency and precise content editing, these methods also suffer from insufficient feature decomposition (implicit editing), leading to artifacts and limited editing capabilities, such as the inability to independently transfer style or background.

\subsection{Explicit Disentanglement Methods}
Explicit feature disentanglement aims to enhance the quality and flexibility of STE by separating visual components.

Early STE research, primarily based on Generative Adversarial Networks (GANs)\cite{gan}, often utilized distinct modules for tasks such as text conversion, background inpainting, style extraction, and foreground-background fusion\cite{stesurvey}. This modular approach represented an early form of explicit disentanglement. Methods like SRNet \cite{srnet}, SwapText \cite{swaptext}, TextStyleBrush \cite{textstylebrush}, and MOSTEL \cite{mostel} focused on tasks such as word or text-line editing, content replacement, and aesthetic transfer. Despite this modularity, their generalization ability was limited by the inherent capacity constraint of GANs and the difficulty of accurately decomposing text styles, leading to unstable background recovery and undesirable fusion artifacts \cite{stesurvey}.

Diffusion models, particularly in the cropping-reconstruction paradigm, also enable explicit disentanglement. SceneVTG \cite{scenevtg} employs MLLM reasoning for text generation, TextCtrl \cite{textctrl} controls through style and glyph disentanglement, and FontDiffuser \cite{fontdiffuser} focuses on one-shot font generation for diverse styles. However, TextCtrl’s binary disentanglement (coupling style and background) often results in stylistic entanglement and unreliable background guidance from unmasked regions.

Our \textbf{TripleFDS} framework advances previous disentanglement methods. Unlike binary disentanglement, which entangles style and background, our approach achieves precise disentanglement of triple features, resulting in truly independent features. This robust disentanglement enables flexible combinations and diverse STE operations, addressing the stylistic entanglement and boundary issues faced by previous methods. 

\begin{figure*}[t]
\centering
\includegraphics[width=0.9\textwidth]{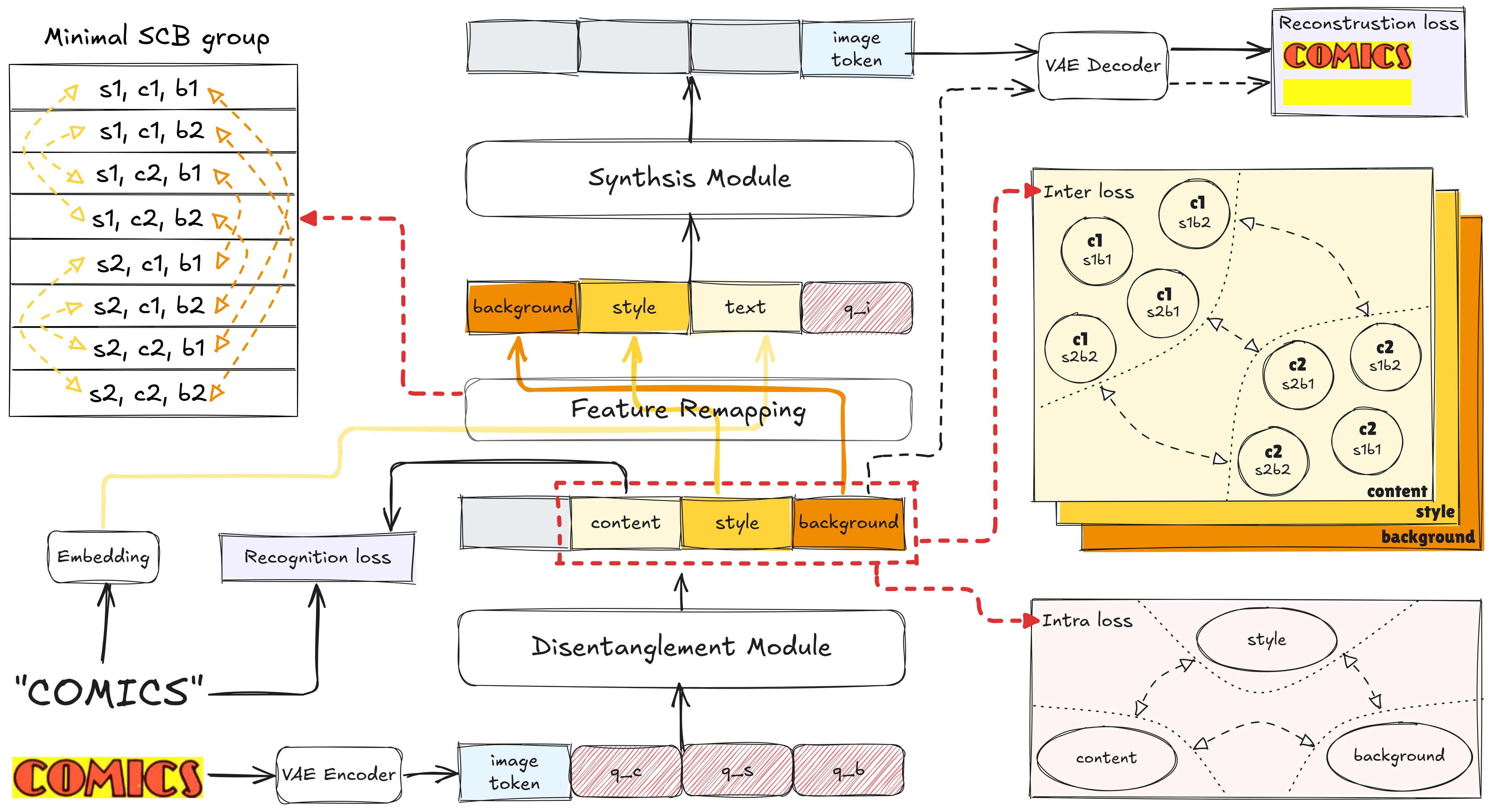}
\caption{Overview of our \textbf{TripleFDS} framework. This figure illustrates its core components and strategies: a minimal SCB Group (top-left) where yellow and brown lines denote remapping objects for style and background features under the \textit{Remapping Strategy}; the pipeline for feature disentanglement and synthesis (middle), with rounded rectangles representing network structures, red diagonal lines indicating learnable token; and visualizations of the \textit{Inter loss} and \textit{Intra loss} (bottom-right).}
\label{fig:pipeline}
\end{figure*}

\section{Method}
This section details our proposed Scene Text Editing (STE) framework, \textbf{TripleFDS}, and its supporting dataset \textbf{SCB Synthesis}. Given that \textbf{TripleFDS} relies on the SCB Group as a fundamental data unit, we first introduce the novel dataset construction paradigm in Section~\ref{sec:scb_dataset_paradigm}. Subsequently, Section~\ref{sec:overview} provides an overview of the \textbf{TripleFDS} framework, explaining its disentanglement and synthesis processes. In Section~\ref{sec:disentanglement_constraints}, we present the key feature disentanglement constraints, which ensure semantic accuracy and non-redundancy of features. Finally, Section~\ref{sec:remapping_strategy} discusses our feature remapping strategy, a crucial component for robust feature learning and preventing model collapse.

\subsection{SCB Dataset Construction Paradigm}
\label{sec:scb_dataset_paradigm}
We introduce a systematic method for constructing text image datasets by explicitly decomposing complex scene text images into three independent dimensions: \textit{background}, \textit{content}, and \textit{style}. Background refers to image regions outside the text mask, content denotes semantic characters within the text mask, and style encompasses visual features inside the text mask, such as font color, texture, glyph, border, and graphic transformations.

As depicted in Fig.~\ref{fig:scb}, we propose the SCB Group concept. An SCB Group is an image collection formed by systematically combining information from these dimensions. An STE image $I$ is modeled as $I = \mathcal{G}(S, C, B)$. For example, a minimal SCB Group comprises 8 text images derived from pairwise combinations of two styles $S_x$, two contents $C_y$, and two backgrounds $B_z$, resulting in $2 \times 2 \times 2 = 8$ variations, expressed as $\{ I_{S_x C_y B_z} \mid x, y, z \in \{1, 2\} \}$. With the SCB Group, we can not only construct editing pairs for various features (e.g., content editing, style switching, and background changing), but also leverage its inherent structure, specifically the fixed feature correspondence among samples, as depicted in Fig.~\ref{fig:pipeline} (top-left). This enables robust training via the feature remapping strategy (Section~\ref{sec:remapping_strategy}) and fosters accurate feature disentanglement through contrastive learning with positive/negative sample pairing (Section~\ref{sec:disentanglement_constraints}).

To mitigate the significant domain gap observed between synthetic and real datasets in prior methods, we synthesize data with enhanced realism. The detailed methodologies for foreground text and style processing, including font selection, color configuration, and graphic transformations, as well as the strategies for foreground and background fusion, are thoroughly described in Appendix~B.

\subsection{Overall Framework}
\label{sec:overview}  
Our framework \textbf{TripleFDS} consists of two key processes: feature disentanglement and feature synthesis. Before describing these processes, we define the key notations used throughout this section: the VAE\cite{vae} sampling rate is denoted as $f$, the hidden dimension is $d$, and the character vocabulary size is $|\Sigma|$. The image embedding sequence length is $N = (H \times W) / f^2$ (where $H$ and $W$ represent the image height and width), and the text embedding sequence length is $L$ (representing the maximum input character count).

During training, we first perform feature disentanglement to extract the core triple features, followed by feature synthesis to reconstruct the image.

\begin{enumerate}
\item \textbf{Feature Disentanglement.} The first process involves disentangling the input image into three key features: content, style, and background. This is achieved using a Transformer-decoder-based disentanglement module $\mathcal{F}_{\text{disentangle}}$ \cite{gpt2}. The input image embedding $E_{I_{\text{src}}} \in \mathbb{R}^{N \times d}$ is obtained via a VAE encoder $\mathcal{E}$, while learnable query tokens ($Q_C \in \mathbb{R}^{L \times d}$, $Q_S \in \mathbb{R}^{(N-L) \times d}$, $Q_B \in \mathbb{R}^{N \times d}$) guide the extraction of the content, style, and background features. The disentanglement operation produces the following features:
\begin{equation}
[E_{\text{ignore}},\! E_{C_{\text{src}}},\! E_{S_{\text{src}}},\! E_{B_{\text{src}}}] \!=\! \mathcal{F}_{\text{disentangle}}([E_{I_{\text{src}}},\! Q_C,\! Q_S,\! Q_B]),
\end{equation}
where $E_{C_{\text{src}}} \in \mathbb{R}^{L \times d}$ represents the content feature, $E_{S_{\text{src}}} \in \mathbb{R}^{(N-L) \times d}$ represents the style feature, and $E_{B_{\text{src}}} \in \mathbb{R}^{N \times d}$ represents the background feature.
 $E_{\text{ignore}}$ serves as a placeholder. These disentangled features are further constrained using the self-supervised regularization loss from Section~\ref{sec:disentanglement_constraints}, ensuring semantic accuracy and non-redundancy.

\item \textbf{Feature Synthesis.} In the second process, the disentangled background and style features ($E_{B_{\text{src}}}$, $E_{S_{\text{src}}}$) are combined with the text embedding $E_{T_{\text{src}}} \in \mathbb{R}^{L \times d}$, derived from a character embedding lookup table $E_{\text{char}} \in \mathbb{R}^{|\Sigma| \times d}$, and an image reconstruction query $Q_I \in \mathbb{R}^{N \times d}$. This process uses another Transformer-decoder-based module $\mathcal{F}_{\text{synth}}$ for feature synthesis. Before synthesis, we apply the feature remapping strategy from Section~\ref{sec:remapping_strategy} to create a ``hard-to-reconstruct" triplet, ensuring robust feature learning. The synthesis operation is defined as:
\begin{equation}
[E_{\text{ignore}}, E_{I_{\text{rec}}}] = \mathcal{F}_{\text{synth}}([E_{B_{\text{src}}}, E_{S_{\text{src}}}, E_{T_{\text{src}}}, Q_I]).
\end{equation}
In this process, $E_{I_{\text{rec}}}$ represents the embedding of the reconstructed image. Notably, instead of using the content feature $E_{C_{\text{src}}}$, the module leverages the text embedding $E_{T_{\text{src}}}$ to guide content generation, ensuring accurate and pure content guidance during both training and inference.
\end{enumerate}

During inference, we apply the same framework as in training, but with modified inputs depending on the task. For text editing, we replace the input text with the target text. For style or background transfer tasks, we use the source image and condition image as input. The relevant feature from the condition image is used to replace the corresponding feature in the source image before the synthesis process.

\subsection{Feature Disentanglement Constraints}
\label{sec:disentanglement_constraints}
Feature disentanglement is crucial for stable reconstruction and robust feature learning, its efficacy depends critically on the purity of the features from the disentanglement process. If the disentanglement process produces under-disentangled or redundant representations, it will severely hinder both the learning process and reconstruction accuracy.

Therefore, precise feature disentanglement is achieved via tailored contrastive loss functions. Firstly, the \textit{inter-group contrastive loss $\mathcal{L}_{\text{inter}}$} is employed to ensure that each feature (background, style, content) accurately corresponds to its respective image component and semantic token, thereby facilitating effective feature disentanglement across different samples. As shown in Fig.~\ref{fig:pipeline} (right-middle), in a minimal $2 \times 2 \times 2$ SCB Group, content feature learning defines positive and negative samples based on shared content. This concept extends across the batch, incorporating samples from other SCB Groups as additional negative samples to increase learning difficulty and foster training stability. High-dimensional features are projected for dimensionality reduction and then normalized to enable cosine similarity computation for contrastive learning, and an improved multi-round InfoNCE\cite{infonce} loss optimizes style, content, and background features by adjusting positive/negative sample partitioning. The InfoNCE loss is:
\begin{equation}
\mathcal{L}_{\text{inter}_{i,j}}
\!=\! - \log \frac{e^{sim(i,j)/\tau}}{e^{sim(i,j)/\tau} \!+\! \sum_{k \notin (\{i\} \cup P_i)} e^{sim(i,k)/\tau} \!+\! \epsilon}. 
\end{equation}

Here, $P_i$ denotes the set of indices of all positive samples for anchor $i$ (excluding $i$ itself). The term $sim(i,j)$ denotes the cosine similarity between projected feature vectors $F_i$ and $F_j$ of anchor $i$ and sample $j$. $\tau$ is a learnable temperature parameter, and $\epsilon$ is a small constant for numerical stability. The final $\mathcal{L}_{\text{inter}}$ loss is computed by averaging all such $\mathcal{L}_{\text{inter}_{i,j}}$ terms across all samples and all triple feature types (background, style, and content), aiming to maximize similarity between anchors and their positive samples while minimizing similarity with all negative samples.

To obtain purer disentangled features and mitigate implicit intra-sample coupling, we employ an \textit{intra-sample multi-feature similarity loss} $\mathcal{L}_{\text{intra}}$. This loss calculates the cosine similarity between the features of the projected background $B_i$, the style $S_i$, and the content $C_i$ for each sample. The similarity loss is defined as:
\begin{equation}
\mathcal{L}_{\text{intra}}
\!=\! \frac{1}{3} \left( |\text{sim}(B_i, S_i)| \!+\! |\text{sim}(B_i, C_i)| \!+\! |\text{sim}(S_i, C_i)| \right).
\end{equation}

Minimizing these similarity scores forces orthogonality in the latent space, thereby eliminating redundancy, as shown in Fig.~\ref{fig:pipeline} (bottom-right).

\subsection{Feature Remapping Strategy}
\label{sec:remapping_strategy}
As described in Section \ref{sec:scb_dataset_paradigm}, the SCB Group facilitates diverse STE tasks by combining styles, contents, and backgrounds. While this enables various feature editing pairs, designing separate tasks for every permutation becomes overly complex. All tasks fundamentally involve feature disentanglement and recombination, differing primarily in how features are grouped. Therefore, instead of traditional editing-centric training \cite{mostel, textctrl, rsste}, we adopt a source image reconstruction approach: disentangling an image into its triple features and reconstructing the original image using them, as shown in Fig.~\ref{fig:pipeline} (middle).

A challenge in reconstruction training is that one feature (e.g., background) may carry redundant information, allowing the model to ``shortcut", leading to collapse where content and style features are minimally learned. To prevent this, we introduce a \textit{feature remapping strategy}. This strategy uses the fixed feature mapping relationships in the SCB Group to remap features during reconstruction, forcing the model to rely on the correct feature combinations.

After decomposing the original image into text style, text content, and background features, remapped feature triplets are created by leveraging SCB Group mappings. The remapping logic is as shown in Fig.~\ref{fig:pipeline} (top-left).

\begin{itemize}
\item The \textbf{background feature} is remapped with another background from the same SCB Group, ensuring background identity while varying content and style (brown dotted line).
\item The \textbf{style feature} is remapped with another style from the same SCB Group, maintaining style identity but varying background and content (yellow dotted line).
\item The \textbf{content feature} is not remapped, as it is derived directly from the input text embedding, ensuring consistency across images with the same text content.
\end{itemize}

If multiple valid remapping samples exist for a feature, their average forms the final remapped feature.

For the background feature, the remapping formula is:
\begin{equation}
\bar{B}_{\text{remap}}
= \frac{1}{(N_S-1)(N_C-1)} \sum_{\substack{s' \in \{1,\dots,N_S\} \\ c' \in \{1,\dots,N_C\} \\ s' \neq s, c' \neq c}} B|I_{s', c', b}.
\end{equation}
Here, $\bar{B}_{\text{remap}}$ is the new background feature for the current sample $I_{s, c, b}$, and $B|I_{s', c', b}$ is the background feature from image $I_{s', c', b}$ in the SCB Group, averaged over styles $s'$ and contents $c'$.

The hybrid triplet is fed into the feature synthesis module, which must reconstruct the original image. This targeted interference forces the module to rely only on the target image features, penalizing redundant encoding (e.g., style/content in the background) by causing reconstruction failure and higher loss. This mechanism encourages the disentanglement module to learn purer features, ensuring that each feature type (background, style) captures only its relevant information, creating ``hard-to-reconstruct" inputs and establishing an information bottleneck that minimizes inter-feature correlation.

In addition to the self-supervised loss discussed in Section~\ref{sec:disentanglement_constraints}, the availability of ground truth labels for content and background features allows us to apply supervised constraints. Following the design of RS-STE \cite{rsste}, we introduce two loss functions: a reconstruction loss $\mathcal{L}_{\text{rec}}$ for the quality of the inpainting background and the reconstruction image and a recognition loss $\mathcal{L}_{\text{reg}}$ for the accuracy of content feature.

The total loss function is then defined as:
\begin{equation}
\mathcal{L}_{\text{total}} = (\mathcal{L}_{\text{rec}} + \mathcal{L}_{\text{reg}}) + \lambda_{\text{inter}}\mathcal{L}_{\text{inter}} + \lambda_{\text{intra}}\mathcal{L}_{\text{intra}}.
\end{equation}
Here, $\mathcal{L}_{\text{rec}}$ and $\mathcal{L}_{\text{reg}}$ are the baseline losses from RS-STE \cite{rsste}, while $\mathcal{L}_{\text{inter}}$ and $\mathcal{L}_{\text{intra}}$ are new loss terms introduced for our model, with their respective settings of weights are provided in Appendix~C.

\begin{figure*}[t]
\centering
\includegraphics[width=0.9\textwidth]{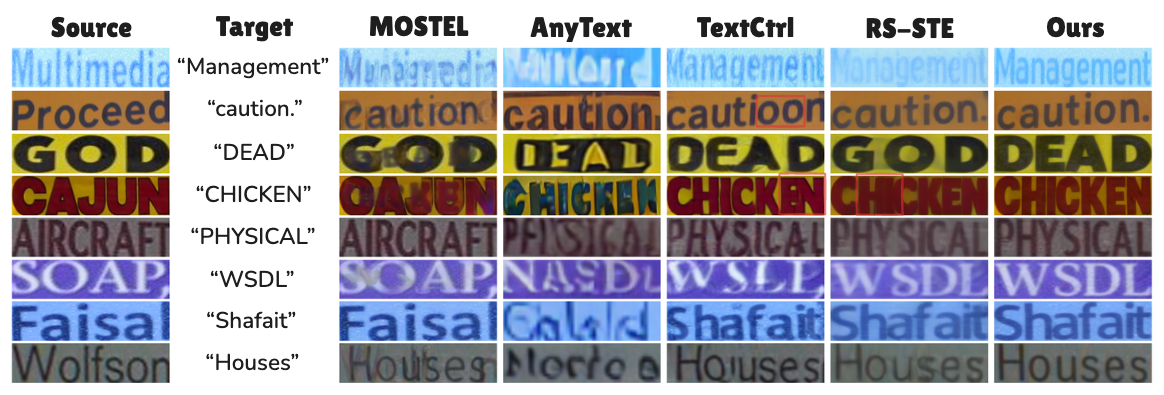}
\caption{Comparison of previous methods with ours. Previous methods tend to generate incorrect or fused text, as shown in the red boxes, while \textbf{TripleFDS} effectively mitigates these problems.}
\label{fig:qualitative}
\end{figure*}

\begin{table*}[t]
\begin{tabular}{l|c|cccc|ccccc}
\toprule
\multirow{2}{*}{Methods} & \multicolumn{1}{c|}{Tamper-Scene} & \multicolumn{4}{c|}{Tamper-Syn2k} & \multicolumn{5}{c}{ScenePair} \\
\cmidrule(lr){2-2} \cmidrule(lr){3-6} \cmidrule(lr){7-11}
& ACC$\uparrow$ & SSIM$\uparrow$ & PSNR$\uparrow$ & MSE$\downarrow$ & FID$\downarrow$ & ACC$\uparrow$ & SSIM$\uparrow$ & PSNR$\uparrow$ & MSE$\downarrow$ & FID$\downarrow$ \\
\midrule
SRNet & 30.26 & 49.97 & 18.66 & 2.16 & 64.37 & 17.84 & 26.66 & 14.08 & 5.61 & 49.22 \\
MOSTEL & 66.54 & 56.94 & 20.27 & 1.35 & 33.79 & 37.69 & 27.45 & 14.46 & 5.19 & 49.19 \\
AnyText & -- & -- & -- & -- & -- & 51.12 & 30.73 & 13.66 & 6.19 & 51.79 \\
DARLING  & 70.85 & 60.07 & 20.80 & 1.20 & 44.48 & -- & -- & -- & -- & -- \\
TextCtrl & 74.17 & 66.60 & 20.79 & 1.30 & 31.13 & 84.67 & 37.56 & 14.99 & 4.47 & \textbf{43.78} \\
RS-STE\dag & 73.71 & 65.91 & 20.92 & \textbf{1.17} & 36.49 & 89.92 & 42.59 & 15.71 & 3.56 & 48.82 \\
\hline
Ours & \textbf{75.62} & \textbf{67.44} & \textbf{21.60} & 1.26 & \textbf{30.84} & \textbf{93.58} & \textbf{44.54} & \textbf{16.53} & \textbf{3.23} & 46.40 \\
\bottomrule
\end{tabular}
\centering
\caption{Comparison on editing performance with previous methods on Tamper-Syn2k, Tamper-Scene and ScenePair. The MSE and SSIM are presented as $(\times10^{-2})$, and RecAcc is presented in percent $(\%)$.}
\label{tab:quantitative}
\end{table*}

\begin{figure*}[t]
\centering
\includegraphics[width=0.9\textwidth]{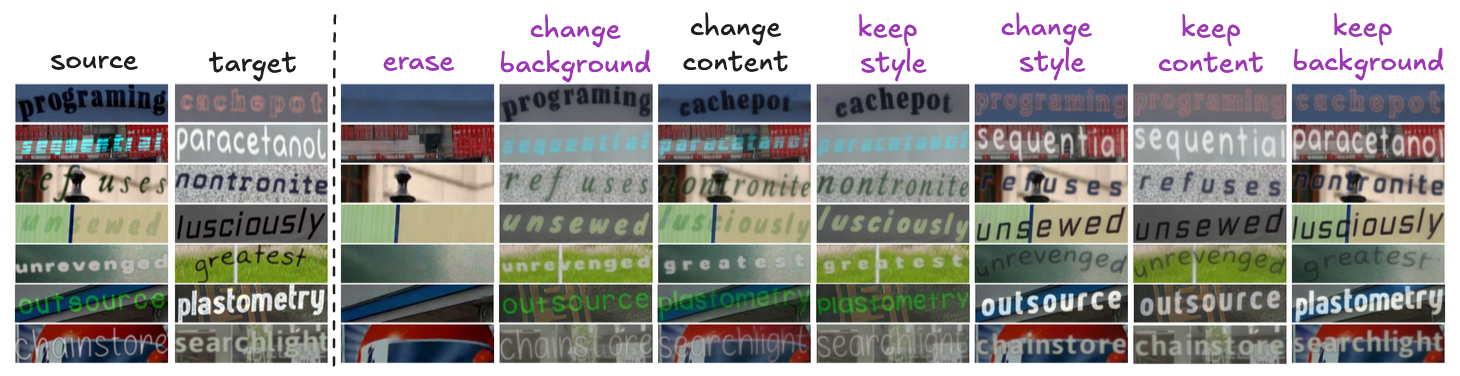}
\caption{Different editing operations of \textbf{TripleFDS}, with the operations highlighted in purple representing those that \textbf{TripleFDS} can perform in addition to the capabilities of previous methods.}
\label{fig:ablation_visualize}
\end{figure*}

\section{Experiments}

\subsection{Datasets and Evaluation Metrics}
\subsubsection{Training Data.}
Our model is trained exclusively on synthetic datasets. During training, we utilized a 1 million-image \textbf{SCB Synthesis} dataset, generated by our dataset construction paradigm (Section~\ref{sec:scb_dataset_paradigm}). For a fair evaluation on Tamper-Syn2k\cite{mostel}, our model was further fine-tuned on MOSTEL's synthetic dataset, Tamper-train-150k.

\subsubsection{Evaluation Data.}
Model performance was comprehensively assessed on various real-world and synthetic datasets, including the ScenePair\cite{textctrl} benchmark (1280 real-world editing-pairs), the Tamper-Syn2k\cite{mostel} dataset (2,000 synthetic editing-pairs), and the Tamper-Scene\cite{mostel} dataset (7,725 unpaired real-world images).

\subsubsection{Evaluation Metrics.}
For visual quality, we adopt: (i) Mean Squared Error (MSE) for pixel difference; (ii) Peak Signal-to-Noise Ratio (PSNR) for signal reconstruction quality; (iii) Structural Similarity Index Measure (SSIM) for structural similarity; and (iv) Fréchet Inception Distance (FID)\cite{fid} for realism and diversity (via feature distribution comparison). For text rendering accuracy, we measure word accuracy (ACC), assessing character-level correctness within edited images.

\subsubsection{Comparison.}
We conduct a quantitative comparison of our \textbf{TripleFDS} against several methods in STE. These include GAN-based (SRNet \cite{srnet}, MOSTEL \cite{mostel}), Diffusion-based (AnyText \cite{anytext}, TextCtrl \cite{textctrl}), and Transformer-based (DARLING \cite{darling}, RS-STE \cite{rsste}) methods. As shown in Tab.~\ref{tab:quantitative}, certain entries are dashed due to specific limitations: AnyText lacks results for Tamper-Syn2k and Tamper-Scene as it is designed for inpainting with cropped inputs. DARLING's ScenePair results are unavailable because its code is not open-source, and its paper only reports performance on Tamper-Syn2k and Tamper-Scene. For a fair comparison, we implement RS-STE (denoted RS-STE\dag{} in Tab.~\ref{tab:quantitative}). This was necessitated by RS-STE's use of MLT2017\cite{mlt2017} for fine-tuning, which could cause data leakage from ScenePair. Our re-implementation, based on their paper and code, ensured fairness by scaling up its parameter count to match ours and adapting our SCB Group dataset for its pretraining.

\subsection{Quantitative and Qualitative Analysis}
As shown in Tab.~\ref{tab:quantitative}, \textbf{TripleFDS} achieves state-of-the-art performance across mainstream STE benchmarks.

For image quality, \textbf{TripleFDS} demonstrates strong performance. Specifically, on Tamper-Syn2k, we achieved leading performance in SSIM (\textbf{67.44}), PSNR (\textbf{21.60}), and FID (\textbf{30.84}). These results indicate that our disentanglement method successfully yields more accurate glyph structures, purer style features, and higher-quality background reconstruction. For MSE, the ``hard-to-reconstruct" triplets employed by the remapping strategy in the synthesis stage, entangled with potential abnormal pixels in synthetic data, can pose challenges to the model, sometimes causing confusion during feature synthesis. On ScenePair, we secured leading performance in SSIM (\textbf{44.54}), PSNR (\textbf{16.53}), and MSE (\textbf{3.23}). \textbf{TripleFDS} achieved the second-best FID of \textbf{46.40}, only behind TextCtrl (FID: \textbf{43.78}), which excels in real-scene background inpainting due to its powerful diffusion architecture. Compared to RS-STE\dag (FID: \textbf{48.82}), \textbf{TripleFDS} significantly improves performance, underscoring the superior effectiveness of explicit disentanglement over implicit editing for high-quality, realistic image generation.

Regarding text accuracy, \textbf{TripleFDS} achieves the highest accuracy on both Tamper-Scene (\textbf{75.62\%}) and ScenePair (\textbf{93.58\%}), surpassing baselines like TextCtrl and RS-STE\dag. Despite RS-STE\dag also employing a recognition loss to emphasize the accuracy of the text, our triple feature disentanglement purifies the style and background features by reducing redundant content information, thus minimizing the appearance of artifacts and improving the accuracy of text recognition.

Upon closer inspection of Fig.~\ref{fig:qualitative}, \textbf{TripleFDS} demonstrates superior editing, effectively avoiding artifacts and color aberrations while maintaining strong style consistency.

\subsection{Ablation}
\label{ablation}
We conducted comprehensive ablation experiments to systematically evaluate the contribution of our key technical designs, hyperparameter choices, and model capacity and training data scale to model performance on the ScenePair dataset.

The following subsections detail experiments on core method designs, SCB Group configurations, and different editing operations.

\textbf{Ablation of core method designs.} As depicted in Tab.~\ref{tab:ablation_method}, our foundational model was trained solely with reconstruction loss $\mathcal{L}_{\text{rec}}$ and recognition loss $\mathcal{L}_{\text{reg}}$.

Introducing the feature remapping strategy RS was crucial, effectively preventing model collapse and improving FID (49.31 to 46.56) and other metrics. This efficacy stems from its implicit diverse feature remapping via ``hard-to-reconstruct" triplets, fully leveraging SCB Group's inherent structure for robust feature learning.

Adding the inter-group contrastive loss \textbf{$\mathcal{L}_{\text{inter}}$} further enhanced performance in ACC, SSIM, and PSNR. Its emphasis on regional feature focus, while maintaining structural integrity, subtly increased FID (46.56 to 47.08) due to residual background feature redundancy interfering with detail recovery.

Finally, incorporating the intra-sample multi-feature similarity loss \textbf{$\mathcal{L}_{\text{intra}}$} completed our full model. This orthogonality constraint directly mitigated redundancy, further improving FID. However, it marginally impacted overall feature expressiveness, resulting in a minor SSIM drop.

\begin{table}[t]
\begin{tabular}{lcccc}
\toprule
Methods & ACC$\uparrow$ & SSIM$\uparrow$ & PSNR$\uparrow$ & FID$\downarrow$ \\
\midrule
$(\mathcal{L}_{\text{rec}} + \mathcal{L}_{\text{reg}})$ & 90.23 & 41.84 & 15.48 & 49.31 \\
+ RS & 92.24 & 43.18 & 16.01 & 46.56 \\
+ $\mathcal{L}_{\text{inter}}$ & 93.14 & \textbf{44.73} & 16.49 & 47.08 \\
+ $\mathcal{L}_{\text{intra}}$ (Ours) & \textbf{93.58} & 44.54 & \textbf{16.53} & \textbf{46.40} \\
\bottomrule
\end{tabular}
\centering
\caption{Ablation results for core method design.}
\label{tab:ablation_method}
\end{table}

\textbf{Ablation of different SCB Group configurations.} This ablation study investigates the impact of SCB Group's internal structure and diversity distribution on disentanglement effectiveness. As detailed in Tab.~\ref{tab:ablation_data_strategies}, we investigated various SCB Group configurations (e.g., (1,4,4,4), (4,4,2,2), (4,2,4,2), (4,2,2,4), (8,2,2,2)), each totaling a 64-image batch, exploring distributions of groups, styles, contents, and backgrounds.

Results show that biasing a single feature dimension degrades overall performance. For example, (4,2,4,2) showed notably worse image quality (FID: 49.48), despite increased content diversity. More balanced configurations like (1,4,4,4) and especially our standard (8,2,2,2) yielded better overall results.

The optimal performance of (8,2,2,2) stems from increased negative sample diversity for contrastive learning, benefiting from a larger number of groups. Conversely, when feature dimensions exceed two (e.g., in (1,4,4,4)), averaging multiple remapping objects in the feature remapping strategy can significantly increase training difficulty due to limited model capacity, thus favoring more groups with fewer elements per dimension.

\begin{table}[t]
\begin{tabular}{lcccc}
\toprule
Configuration & ACC$\uparrow$ & SSIM$\uparrow$ & PSNR$\uparrow$ & FID$\downarrow$ \\
\midrule
(1,4,4,4) & 93.13 & 43.53 & 15.86 & 47.52 \\ 
(4,4,2,2) & 90.89 & 42.17 & 15.76 & 47.59 \\ 
(4,2,4,2) & 92.2 & 41.95 & 15.63 & 49.48 \\ 
(4,2,2,4) & 90.56 & 42.63 & 15.7 & 46.79 \\ 
(8,2,2,2) & \textbf{93.58} & \textbf{44.54} & \textbf{16.53} & \textbf{46.40} \\ 
\bottomrule
\end{tabular}
\centering
\caption{Ablation results for training data configuration. (Number of groups, styles per group, content per group, background per group) indicating image quantity combination, totaling 64 images per batch.}
\label{tab:ablation_data_strategies}
\end{table}

\textbf{Qualitative Analysis of Various Text Editing Operations.} Fig.~\ref{fig:ablation_visualize} presents the visualization results of \textbf{TripleFDS} across various editing operations. \textbf{TripleFDS} successfully disentangles text, style, and background, offering a wider array of possibilities for scene text editing.

\section{Conclusion}
To overcome challenges in feature disentanglement in prior scene text editing (STE) methods, we introduce \textbf{TripleFDS}, a novel framework with disentangled modular attributes, alongside the \textbf{SCB Synthesis} dataset. This dataset provides robust training data for triple feature disentanglement using the SCB Group. Leveraging this construct as a foundational unit, \textbf{TripleFDS} first disentangles the triple features, ensuring semantic accuracy by applying inter-group contrastive regularization and minimizing feature redundancy. During synthesis, \textbf{TripleFDS} remaps features to prevent ``shortcut" phenomena and avoid feature leakage in reconstruction. Extensive experiments show that \textbf{TripleFDS} outperforms existing methods, offering enhanced flexibility and high-quality results, achieving sota performance in STE.

\section{Acknowledgments}
This work is supported in part by National Natural Science Foundation of China (Grant No. 62276121), the TianYuan funds for Mathematics of the National Science Foundation of China (Grant No. 12326604).

\bibliography{aaai2026}

\newpage
\appendix

\section{Implementation Details}
\label{implementation}
The core of our framework consists of two parallel Transformer decoders (GPT-2 \cite{gpt2} style, hidden dimension 384, 12 layers, 6 attention heads, each with 64-dimensional vectors), utilized in both the disentanglement and synthesis phases, with a total of 50M trainable parameters. The image encoder and decoder use an f8-KL VAE \cite{vae}, where f8 refers to the sampling rate. It is derived from the Latent Diffusion Model \cite{stablediffusion} and was chosen for its superior FID compared to VQ-VAE, as demonstrated in the ablation studies of RS-STE \cite{rsste}.

A text embedding layer encodes the target text using a 95-character set (letters, numbers, symbols, and space) and special tokens, totaling 114 unique tokens. The maximum text sequence length is 16 (with padding), the style token sequence length is 240, and the image/background latent sequence length is 256. The total concatenated sequence length for both phases is 776, with input/output image dimensions of $256 \times 64$ pixels.

Training used a batch size of 64, with batches composed of images from eight $2 \times 2 \times 2$ SCB Groups. The learning rate was $1 \times 10^{-4}$, with $\lambda_{\text{inter}}$ fixed at 0.5 and $\lambda_{\text{intra}}$ adaptively adjusted, typically around 0.25 during training.

\begin{figure*}[t]
\centering
\includegraphics[width=1.0\textwidth]{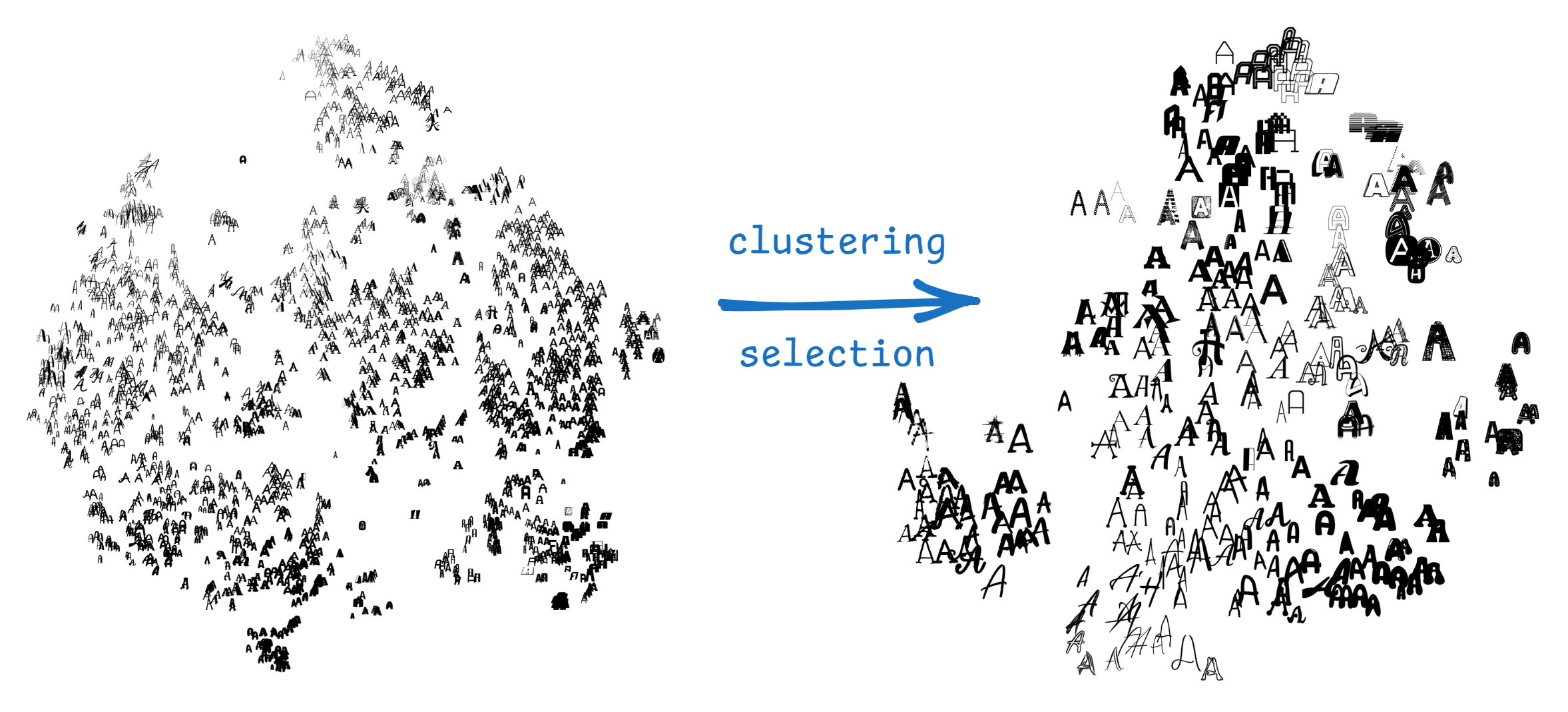} 
\caption{Visualization of font glyph features before (left) and after (right) clustering selection. Each glyph represents the character "A" rendered from a different font. Clustering reduces redundancy while maintaining visual diversity.}
\label{fig:clustering}
\end{figure*}

\begin{figure*}[t]
\centering
\includegraphics[width=1.0\textwidth]{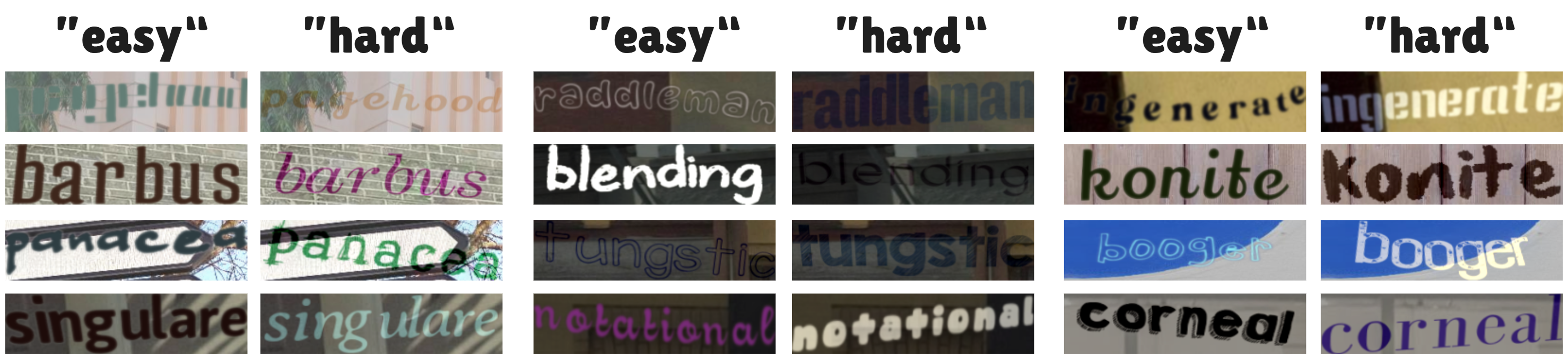} 
\caption{Examples of ``easy" and ``hard" foreground-background fusion samples used during dataset generation. ``Hard" samples exhibit stronger background interference, while ``easy" samples preserve clearer font boundaries.}
\label{fig:fusion_examples}
\end{figure*}

\begin{figure}[t]
\centering
\includegraphics[width=\linewidth]{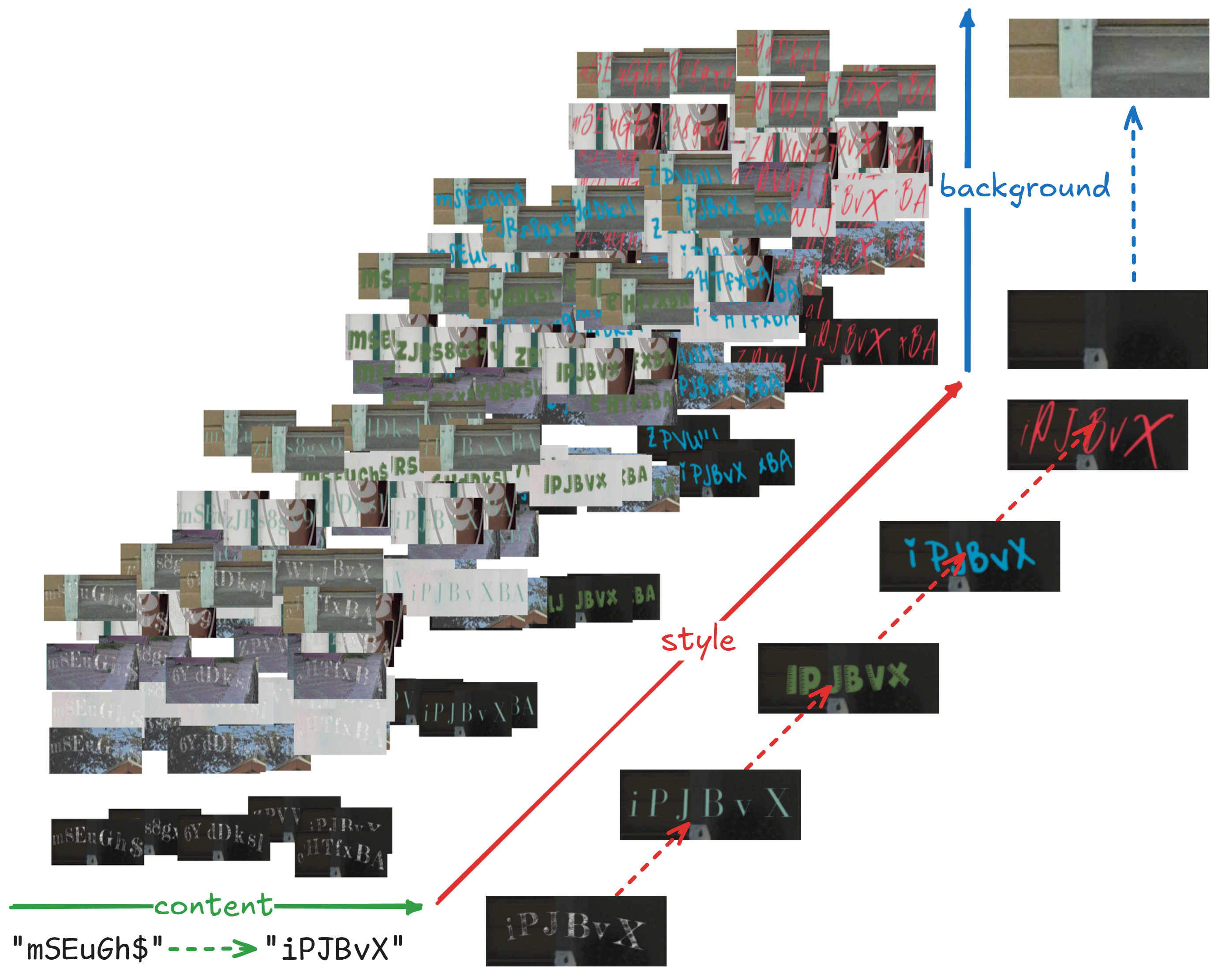} 
\caption{T-SNE-based 3d visualization of disentangled features.}
\label{fig:tsne}
\end{figure}

\begin{figure*}[t]
\centering
\includegraphics[width=1.0\textwidth]{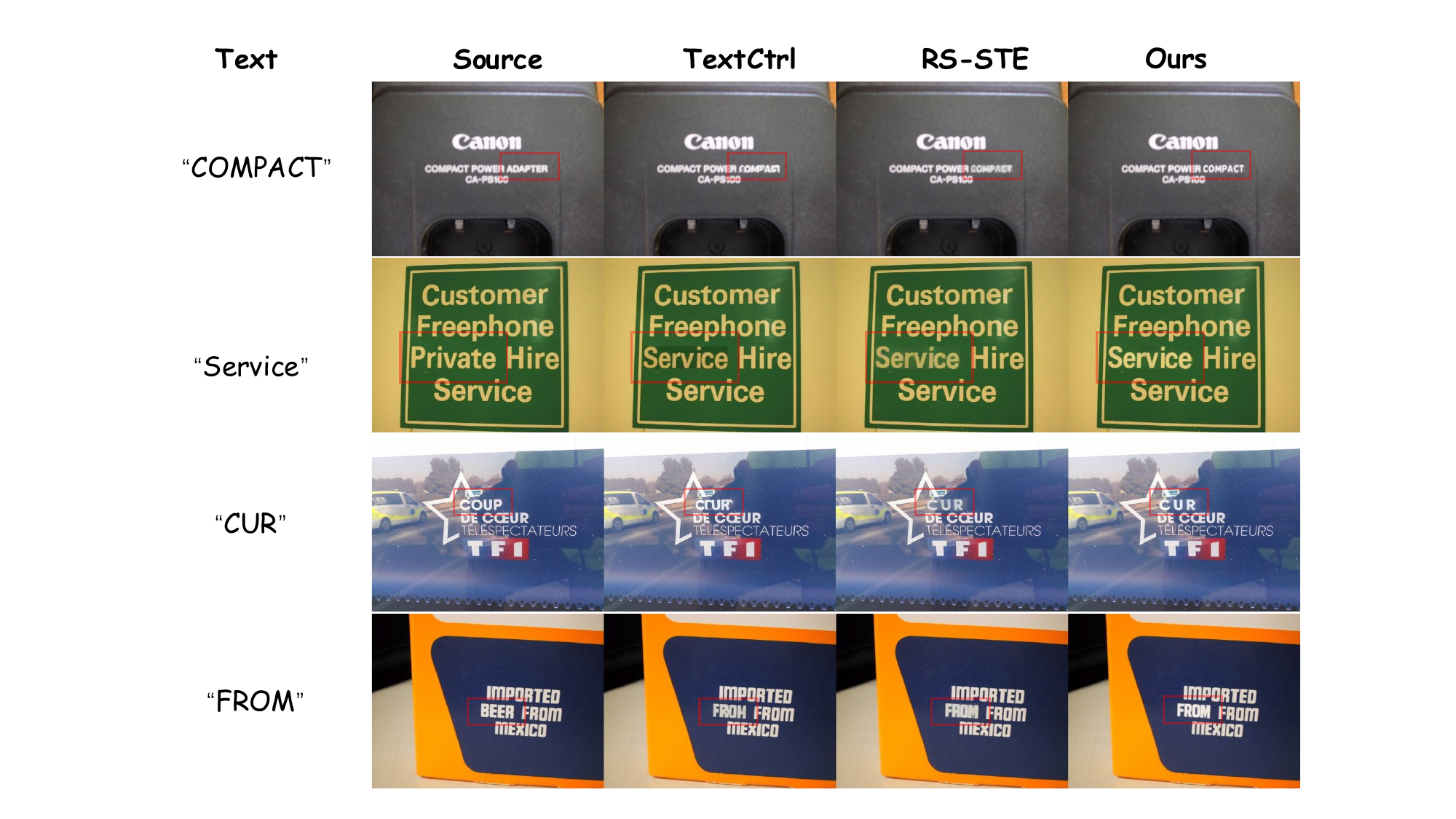} 
\caption{Qualitative comparison on full-sized images. The editing region is marked by the red box. As shown, \textbf{TripleFDS} demonstrates strong text accuracy and effective background restoration.}
\label{fig:full_image}
\end{figure*}

\begin{figure*}[t]
\centering
\includegraphics[width=1.0\textwidth]{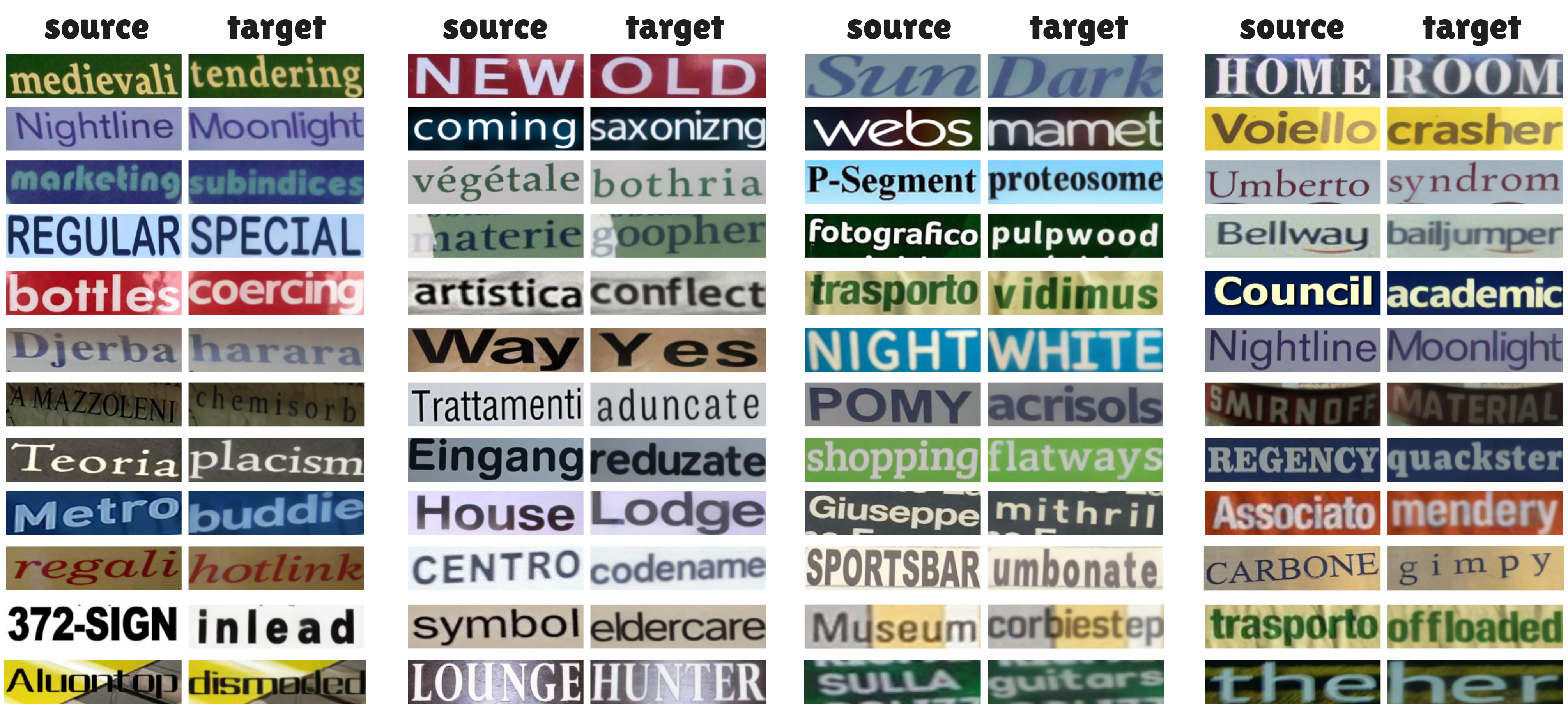} 
\caption{Additional visualization results edited by \textbf{TripleFDS} on the real-world dataset Tamper-Scene.}
\label{fig:tamper-scene}
\end{figure*}

\section{Dataset Construction and Processing}

This research provides a detailed expansion of the SCB Dataset Construction Paradigm, which systematically constructs text image datasets by explicitly disentangling complex scene text images into three independent dimensions. For background materials, we adopt the SceneVTG-Erase\cite{scenevtg} pure background image set as our data source. This dataset provides clean and diverse backgrounds without text, offering richer textures and structural complexity compared to simple backgrounds. It challenges feature disentanglement and better aligns the training environment with real-world scenarios.

\subsection{Foreground Text and Style Processing}

Preparing foreground text (content) and style is essential for dataset construction. For text content, we use character-level encoding, enabling random character assembly. To ensure training stability and simulate real-world language distributions, a weighted character selection mechanism is implemented (\textit{e.g.,} $letters > numbers > symbols$). This sampling approach reduces cluttered typesetting. Text lengths range from 3 to 14 characters, with a maximum difference of 3 characters within a group, following a peaked probability distribution.

Font style processing refines several key aspects:

\begin{itemize}
\item \textbf{Font Selection and Clustering Optimization.} We collected approximately 3000 candidate fonts. To build a compact and representative library, we rendered 95 standard character images ($0$--$9$, $a$--$z$ in both cases, and punctuation) for each font and resized them to $128 \times 128$ pixels. Glyph features were extracted using ResNet-50 \cite{resnet}, followed by clustering to identify 500 representative font families. The closest fonts to each cluster center were retained for training, ensuring diversity while minimizing redundancy. Fig.~\ref{fig:clustering} illustrates the glyph distribution before and after clustering.
\item \textbf{Font Color Configuration.} Colors are sampled in RGB, with practical shades (black, white, gray) covering over 50\% of the palette. Generated font colors are validated to ensure sufficient contrast with the background and distinction from other group colors, enhancing readability and diversity.
\item \textbf{Font Graphic Transformation Enhancement.} Based on SRNet-Datagen \cite{srnet}, we increased the probability and range of text bending, and expanded the scope for rotation, shearing, and scaling. These augmentations improve text image diversity and complexity, facilitating robust text representation learning.
\end{itemize}

During foreground synthesis, maintaining font typesetting and text region consistency is crucial. The objective is to preserve the coherence of the text region before and after editing. For most cropped text images, the aspect ratio is approximately 4. We set the foreground size to $256 \times 128$ pixels, which is sufficient to accommodate reasonably tilted and curved text. In a minimal group, selected content and styles generate foregrounds, ensuring that the max-scaled foreground closely fits the width/height limits for a given font, with varying content. The background cropping size is determined by the maximum width/height across content/style variations. Adaptive character spacing is applied to handle changes in character count, ensuring consistent text region dimensions and minimizing distortion, thereby simulating realistic editing.

\subsection{Foreground and Background Fusion}
\label{sec:fusion}
Foreground and background fusion plays a critical role in both realism and training complexity. We randomly select two dataset images, cropping predefined regions as backgrounds. The centered foreground is then fused with the background. The fusion difficulty is categorized as follows:
\begin{itemize}
\item \textbf{``Hard" Level.} Gradient-based Poisson blending is used for smooth transitions. In complex backgrounds with large color differences, it may affect font brightness and distinctness, increasing the challenge for model disentanglement.
\item \textbf{``Easy" Level.} Simple foreground-background splicing with basic edge processing, resulting in clearer boundaries and lower recognition difficulty.
\end{itemize}
To improve robustness, we use a balanced training strategy, where ``easy" and ``hard" fusion methods are applied with equal probability during dataset generation, referred to as ``medium". The corresponding ablation study is detailed in Tab.~\ref{tab:ablation_fusion_strategies}. Examples of ``easy" and ``hard" samples are shown in Fig.~\ref{fig:fusion_examples}, highlighting differences in contrast, background complexity, and blending artifacts.E

To address the over-brightness and excessive clarity of synthesized data (compared to real-world images), we apply visual adjustments: brightness and contrast are randomly perturbed, and Gaussian blur is optionally applied. These augmentations simulate realistic lighting and imaging conditions, enhancing the diversity and realism of the dataset.

\section{More Experiments}
\textbf{Ablation of Inter-Group Contrastive Loss Weight ($\lambda_{\text{inter}}$).} We conducted an ablation study to determine the optimal weight coefficient for the inter-group contrastive loss ($\lambda_{\text{inter}}$). For $\mathcal{L}_{\text{intra}}$, its weight was dynamically adjusted during training, inspired by Stable Diffusion's VAE training \cite{stablediffusion}. This dynamic adjustment employed an Exponential Moving Average (EMA) update based on the gradient ratio of the projection head's ($\mathcal{P}$) final linear layer to balance gradient influence.

The optimization for $\lambda_{\text{inter}}$ involved two phases. A coarse search (0, 0.1, 1.0, 10.0) identified $\lambda_{\text{inter}}=1.0$ as a robust performance point. Subsequently, a finer-grained search (0.25, 0.5, 1.0, 2.0) around this range, as shown in Tab.~\ref{tab:ablation_lambda_inter}, revealed $\lambda_{\text{inter}}=0.5$ as optimal. This value offers the best balance across image quality metrics (SSIM, PSNR, FID), despite a slightly lower ACC than $\lambda_{\text{inter}}=0.1$, and was adopted for our full model.

We further select $\lambda_{\text{inter}}$ on a unseen 10k SCB validation set, judged by $\mathcal{L}_{\text{rec}}+\mathcal{L}_{\text{reg}}$. The optimal value of 0.25, while different from original choice, still yielded SOTA result on ScenePair (Tab.~\ref{tab:ablation_lambda_inter}), confirming method was not over-tuned to the test metrics.

\begin{table}[t]
\centering
\begin{tabular}{lcccc}
\toprule
$\lambda_{\text{inter}}$ & ACC$\uparrow$ & SSIM$\uparrow$ & PSNR$\uparrow$ & FID$\downarrow$ \\
\midrule
0.0 & 92.24 & 43.18 & 16.01 & 46.56 \\
0.1 & \textbf{93.75} & 44.01 & 16.31 & 46.59 \\
0.25 & \underline{93.61} & \underline{44.35} & 16.26 & \underline{46.51} \\
0.5 & 93.58 & \textbf{44.54} & \textbf{16.53} & \textbf{46.40} \\
1.0 & 93.05 & 44.21 & \underline{16.37} & 46.86 \\
2.0 & 90.75 & 42.94 & 16.28 & 50.23 \\
10.0 & 74.70 & 37.23 & 14.74 & 64.49 \\
\bottomrule
\end{tabular}
\caption{Ablation results for the weight coefficient of inter-group contrastive loss ($\lambda_{\text{inter}}$). \textbf{Bold} values indicate the best performance, while \underline{underlined} values represent the second-best results.}
\label{tab:ablation_lambda_inter}
\end{table}

\textbf{Ablation of Foreground-Background Fusion Strategies.}
We evaluate the impact of different fusion strategies introduced in Section~\ref{sec:fusion}, including ``Easy," ``Hard," and the ``Medium" mix. As shown in Tab.~\ref{tab:ablation_fusion_strategies}, the ``Easy" configuration achieves the highest ACC and PSNR due to clearer font boundaries, while the ``Hard" setting yields the best FID, benefiting from realistic blending. The ``Medium" strategy balances both, achieving the highest SSIM and strong overall performance, validating its effectiveness in training under diverse conditions.

\begin{table}[t]
\begin{tabular}{lcccc}
\toprule
Strategies & ACC$\uparrow$ & SSIM$\uparrow$ & PSNR$\uparrow$ & FID$\downarrow$ \\
\midrule
Easy & \textbf{94.03} & \underline{44.38} & \textbf{16.74} & 50.39 \\
Medium & \underline{93.58} & \textbf{44.54} & \underline{16.53} & \underline{46.40} \\
Hard & 91.33 & 42.15 & 16.09 & \textbf{45.95} \\
\bottomrule
\end{tabular}
\centering
\caption{Ablation results for different foreground and background fusion strategies.}
\label{tab:ablation_fusion_strategies}
\end{table}

\textbf{Ablation of Image Sequence Length.}
We investigate the impact of image token sequence length on performance. Given an image size of $256 \times 64$, using a VAE with a downsampling factor $f=8$ results in a latent resolution of $32 \times 8 = 256$ tokens, while using $f=4$ yields a latent resolution of $64 \times 16 = 1024$ tokens—4$\times$ longer than the former. As shown in Tab.~\ref{tab:ablation_seqlen}, a lower downsampling rate ($f=4$) achieves better visual quality, as reflected by improved FID. However, the significantly longer image token sequence dilutes the influence of style and content tokens during attention, weakening control over textual structure. As a result, metrics such as SSIM and ACC slightly degrade compared to the $f=8$ (256-token) setting.

\begin{table}[t]
\begin{tabular}{lcccc}
\toprule
Sequence Length & ACC$\uparrow$ & SSIM$\uparrow$ & PSNR$\uparrow$ & FID$\downarrow$ \\
\midrule
256 & \textbf{93.58} & \textbf{44.54} & \textbf{16.53} & 46.40 \\
1024 & 90.91 & 43.86 & 16.35 & \textbf{41.24} \\
\bottomrule
\end{tabular}
\centering
\caption{Ablation results for different image sequence length.}
\label{tab:ablation_seqlen}
\end{table}

\textbf{The Significance of the SCB Dataset.} 
Our dataset's SCB Group structure fundamentally differs from the editing-pair paradigm of prior work. This structure enables a reconstruction-based training objective, which demonstrates clear advantages over traditional pair-based training (Tab.~\ref{tab:ablation_method}, row 2 vs. row 1).

To further validate our dataset's quality, we retrained RS-STE\cite{rsste} on our SCB dataset (reformatted as editing-pairs). As detailed in Tab.~\ref{tab:scb_significance}, this model significantly outperformed the original model trained on the same-sized Tamper-train-150k dataset\cite{mostel}. Specifically, when evaluated on ScenePair, our dataset yielded a 15.39\% increase in accuracy, an SSIM improvement of 7.98, and a reduction in FID (from 65.28 to 62.09). These results confirm that our SCB dataset enhances the model's ability to generate high-quality images.

\begin{table}[t]
\begin{tabular}{lcccc}
\toprule
Datasets & ACC$\uparrow$ & SSIM$\uparrow$ & PSNR$\uparrow$ & FID$\downarrow$ \\
\midrule
Tamper-train-150k & 71.95 & 32.11 & 15.55 & 65.28 \\
SCB-editing-pairs & \textbf{87.34} & \textbf{40.09} & \textbf{16.02} & \textbf{62.09} \\
\bottomrule
\end{tabular}
\centering
\caption{Performance comparison of RS-STE trained on Tamper-train-150k and our SCB-editing-pairs.}
\label{tab:scb_significance}
\end{table}

\textbf{Ablation of Model Parameter Count.}
We investigate the impact of model capacity on performance by varying the hidden dimension of our Transformer decoders. Two configurations were evaluated: 50M parameters (dim 384, our default setting) and 180M parameters (dim 768). As shown in Tab.~\ref{tab:ablation_params}, increasing the parameter count does not lead to significant improvements in font structure consistency, suggesting that the current training data may have reached its performance limit. However, a notable improvement is observed in FID, indicating that the increase in parameters enhances the model's ability to reconstruct background details, leading to better robustness in handling complex backgrounds.

\begin{table*}[t]

\begin{tabular}{l|c|cccc|ccccc}
\toprule
\multirow{2}{*}{Parameter Count} & \multicolumn{1}{c|}{Tamper-Scene} & \multicolumn{4}{c|}{Tamper-Syn2k} & \multicolumn{5}{c}{ScenePair} \\
\cmidrule(lr){2-2} \cmidrule(lr){3-6} \cmidrule(lr){7-11}
& ACC$\uparrow$ & SSIM$\uparrow$ & PSNR$\uparrow$ & MSE$\downarrow$ & FID$\downarrow$ & ACC$\uparrow$ & SSIM$\uparrow$ & PSNR$\uparrow$ & MSE$\downarrow$ & FID$\downarrow$ \\
\midrule
50M (dim 384) & \textbf{75.62} & 67.44 & \textbf{21.60} & \textbf{1.26} & 30.84 & \textbf{93.58} & 44.54 & \textbf{16.53} & 3.23 & 46.40 \\
180M (dim 768) & 75.18 & \textbf{68.4} & 19.62 & 1.53 & \textbf{27.6} & 92.88 & \textbf{44.9} & 16.52 & \textbf{3.13} & \textbf{42.66} \\
\bottomrule
\end{tabular}
\centering
\caption{Ablation results for different model parameter counts across Tamper-Syn2k\cite{mostel}, Tamper-Scene, and ScenePair\cite{textctrl} datasets.}
\label{tab:ablation_params}
\end{table*}

\begin{table*}[t]
\centering
\begin{tabular}{l|c|cccc|ccccc}
\toprule
\multirow{2}{*}{Dataset Size} & \multicolumn{1}{c|}{Tamper-Scene} & \multicolumn{4}{c|}{Tamper-Syn2k} & \multicolumn{5}{c}{ScenePair} \\
\cmidrule(lr){2-2} \cmidrule(lr){3-6} \cmidrule(lr){7-11}
& ACC$\uparrow$ & SSIM$\uparrow$ & PSNR$\uparrow$ & MSE$\downarrow$ & FID$\downarrow$ & ACC$\uparrow$ & SSIM$\uparrow$ & PSNR$\uparrow$ & MSE$\downarrow$ & FID$\downarrow$ \\
\midrule
200k & 68.16 & 57.89 & 18.44 & 1.6 & 51.37 & 89.38 & 41.49 & 16.03 & 3.65 & 58.84 \\
500k & 72.54 & 60.78 & 19.38 & 1.64 & 48.92 & 92.19 & 43.36 & 16.08 & 3.52 & 52.19 \\
1M & \textbf{75.62} & \textbf{67.44} & \textbf{21.60} & \textbf{1.26} & \underline{30.84} & \textbf{93.58} & \textbf{44.54} & \textbf{16.53} & \textbf{3.23} & \underline{46.40} \\
2M & \underline{73.79} & \underline{66.61} & \underline{19.99} & \underline{1.47} & \textbf{28.58} & \underline{93.29} & \underline{44.11} & \underline{16.36} & \underline{3.27} & \textbf{42.45} \\
\bottomrule
\end{tabular}
\caption{Ablation results for different training dataset sizes across Tamper-Syn2k, Tamper-Scene, and ScenePair datasets.}
\label{tab:ablation_dataset_size}
\end{table*}

\textbf{Ablation of Training Dataset Size.}
We investigate how training dataset size affects model performance and generalization. Four dataset sizes—200k, 500k, 1M, and 2M images—were evaluated, all constructed via our proposed SCB paradigm. As shown in Tab.~\ref{tab:ablation_dataset_size}, performance improves steadily from 200k to 1M.

However, scaling the training set to 2M yields diminishing returns. While FID continues to improve—dropping from 30.84 to 28.58 on Tamper-Syn2k and from 46.40 to 42.45 on ScenePair—other metrics such as ACC and SSIM plateau or slightly decline. For example, Tamper-Scene ACC drops from 75.62 to 73.79, and similar stagnation is observed across ScenePair and Tamper-Syn2k. This indicates that more training data does not always lead to better overall performance.

We attribute this to domain gap issues. Although Tamper-Syn2k is also synthetic, differences in synthesis pipelines—\textit{e.g., background sources, fonts, rendering}—can cause significant distribution shifts across datasets. As training size increases, the model may overfit to patterns specific to the \textbf{SCB Synthesis} domain, limiting its generalization to other synthetic and real-world datasets. Nevertheless, the consistent FID gains suggest that visual quality, especially for backgrounds, still benefits from exposure to more diverse training textures and patterns.

\section{Visualization}

\textbf{TripleFDS} enables explicit disentanglement of background, style, and content features, going beyond traditional style-preserving content editing limitations. As illustrated in Fig.~\ref{fig:tsne}, we project each image’s content, style, and background features into 1D spaces using t-SNE and combine them into 3D coordinates for visualization. Each point is rendered with its corresponding image, producing a feature-driven layout.

The resulting visualization reveals clear clustering patterns: samples with similar contents, styles, or backgrounds are positioned closely, while those with distinct attributes are separated across axes. Along the content axis (green), the textual content transitions smoothly from ``mSEuGh'' on the left to ``iPJBvX'' on the right, reflecting semantic interpolation. Along the background axis (blue, bottom to top), the backgrounds vary consistently in texture and scene structure. Along the style axis (red), images on the far right and bottom demonstrate a wide spectrum of style variations—including font, color, and glyph—applied to fixed content and background. This spatial structure highlights the effectiveness of our disentanglement, with consistent semantic grouping and orthogonal separation of the three attributes.

For full-sized image editing, we crop the target text region, apply the edits, and paste it back. As shown in Fig.~\ref{fig:full_image}, \textbf{TripleFDS} effectively preserves both text accuracy and background restoration.

We show more results on the real-world Tamper-Scene\cite{mostel} dataset, which contains a variety of font types, transformations, and background styles. As demonstrated in the Fig.~\ref{fig:tamper-scene}, \textbf{TripleFDS} effectively handles complex real-world scenarios, seamlessly editing text while maintaining high-quality background restoration.

\end{document}